\documentclass[letterpaper, 10 pt, conference]{ieeeconf}  
\pdfoutput=1
\usepackage{booktabs}
\usepackage{graphicx}
\usepackage{multirow}
\usepackage{bbding}
\usepackage{color}
\usepackage{url}
\usepackage{amsmath,amssymb}

\usepackage[colorlinks,linkcolor=red,anchorcolor=blue,citecolor=green]{hyperref}

\IEEEoverridecommandlockouts                              

\title{\LARGE \bf
4DRaL: Bridging 4D Radar with LiDAR for Place Recognition using  Knowledge Distillation
}

\author{Ningyuan Huang$^{\dagger}$, Zhiheng Li$^{\dagger}$, and Zheng Fang*
\thanks{$^{\dagger}$These authors contributed equally to this work.} 
\thanks{This work was supported by  the Key Technologies and Applications of Live-Line Fault Elimination Robots for Distribution Networks in Complex Terrain Environments under Grant 5400-202328557A-3-2-ZN. (Corresponding author: Zheng Fang, e-mail: fangzheng@mail.neu.edu.cn)}
\thanks{The authors are all with the Faculty of Robot Science and Engineering, Northeastern University, Shenyang, China.}
}

\begin{document}

\maketitle

\begin{abstract}
Place recognition is crucial for loop closure detection and global localization in robotics. Although mainstream algorithms typically rely on cameras and LiDAR, these sensors are susceptible to adverse weather conditions.
Fortunately, the recently developed 4D millimeter-wave radar (4D radar) offers a promising solution for all-weather place recognition. However, the inherent noise and sparsity in 4D radar data significantly limit its performance. 
Thus, in this paper, we propose a novel framework called 4DRaL that leverages knowledge distillation (KD) to enhance the place recognition performance of 4D radar.
Its core is to adopt a high-performance LiDAR-to-LiDAR (L2L) place recognition model as a teacher to guide the training of a 4D radar-to-4D radar (R2R) place recognition model.
4DRaL comprises three key KD modules: a local image enhancement module to handle the sparsity of raw 4D radar points, a feature distribution distillation module that ensures the student model generates more discriminative features, and a response distillation module to maintain consistency in feature space between the teacher and student models. 
More importantly, 4DRaL can also be trained for 4D radar-to-LiDAR (R2L) place recognition through different module configurations.
Experimental results prove that 4DRaL achieves state-of-the-art performance in both R2R and R2L tasks regardless of normal or adverse weather. 
\end{abstract}

\section{INTRODUCTION}

Place recognition aims to identify previously visited places using current sensor data and match them with a pre-built database. Camera and LiDAR, as the most commonly used sensors for place recognition, have made significant progress in the past decade~\cite{vpr_survey,netvlad, ptcnet}. However, these optical-based sensors still have fundamental drawbacks. Specifically, vision-based algorithms are susceptible to variations of lighting conditions, while LiDAR-based methods are limited by adverse weather conditions (e.g., rain, snow and fog). 

Fortunately, millimeter-wave radar can handle these problems with its longer wavelength signal. Some studies have attempted to exploit a mechanical scanning radar for place recognition~\cite{raplace,kidnapped,Off-the-radar}. Nevertheless, the scanning radar is bulky and expensive, which limits its application scope. 
AutoPlace~\cite{autoplace} first demonstrates the place recognition capability of a 3D single-chip automotive radar, but the low resolution of 3D radar limits its performance.
As the latest advancement in radar sensors, 4D radar not only improves the resolution of the point cloud against 3D radar, but also incorporates height information. Thus, several works have explored the use of 4D radar for robot perception~\cite{4DRadarDataset, vod, transloc4d}. TransLoc4D~\cite{transloc4d} presents the first network for 4D radar place recognition, achieving notable results by using multimodal information (geometry, intensity and velocity) of 4D radar. 
Yet, the noise and sparse nature of radar data still 
hinder its performance.
\begin{figure}[t]
	\centering
        \includegraphics[width=0.95\linewidth]{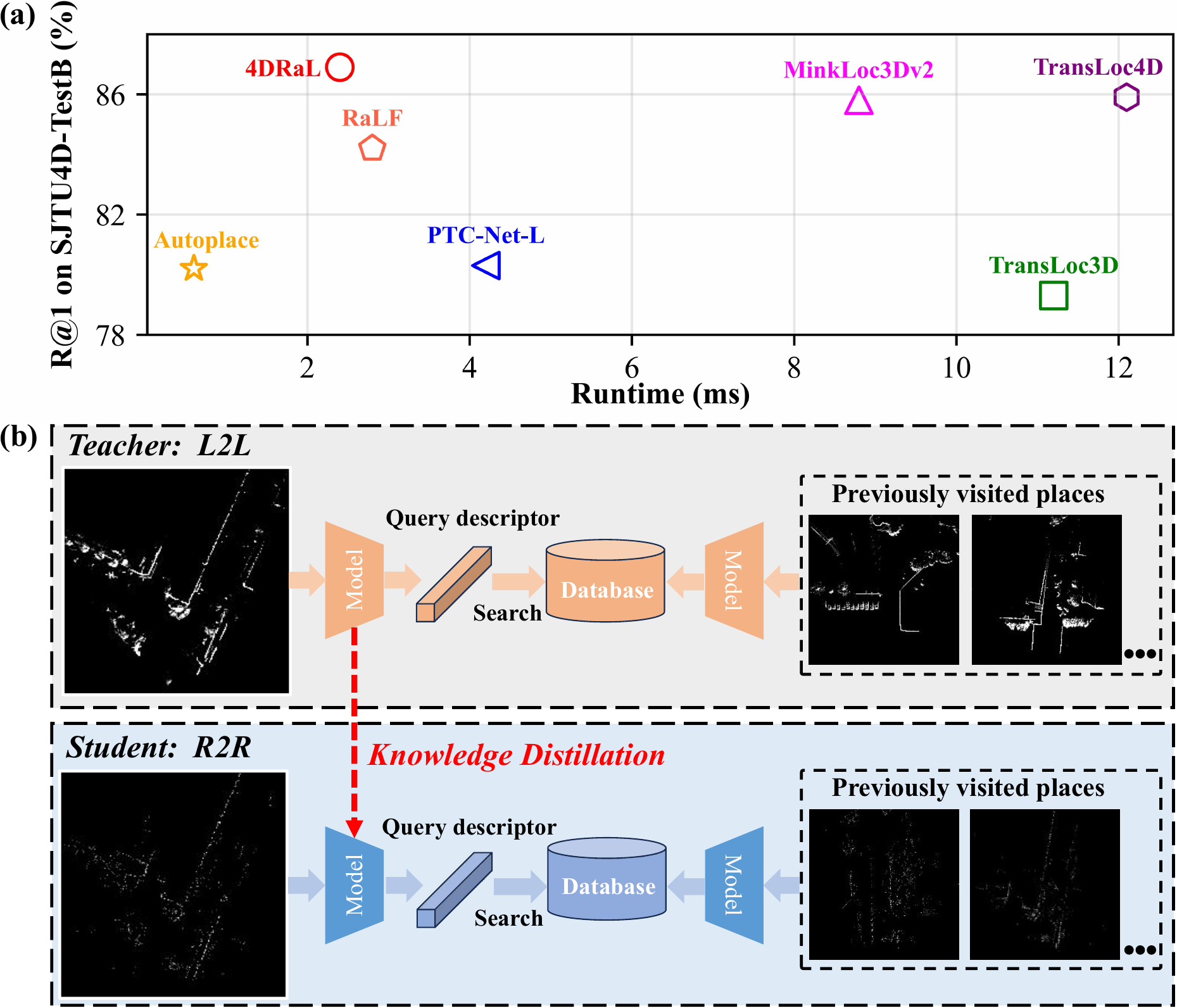}\vspace{-0.3cm}
	\caption{(a) Comparison of runtime and performance across several methods on SJTU4D dataset~\cite{4DRadarDataset}. (b) The framework of 4D radar place recognition based on knowledge distillation. The query scan is encoded into a descriptor, and the database is searched for the closest matching descriptors. The R2R (radar-to-radar) and L2L (LiDAR-to-LiDAR) databases consist of 4D radar descriptors and LiDAR descriptors, respectively.}
	\label{fig::fig1}
         \vspace{-0.6cm}
\end{figure}

To overcome the limitations of 4D radar data, we propose 4DRaL, a novel framework that adopts knowledge distillation (KD) to improve the performance of 4D radar-based place recognition. 
We observe that LiDAR-based place recognition methods generally perform well, largely because LiDAR can provide richer and more accurate perception data compared to 4D radar in favorable weather conditions, allowing it to generate discriminative features. 
Inspired by this, we apply a high-performance LiDAR-to-LiDAR (L2L) place recognition model as a teacher to guide the training process of a radar-to-radar (R2R) place recognition model (Fig.~\ref{fig::fig1}(b)). We aim to enable the 4D radar to generate robust features comparable to LiDAR through KD. In addition, through a streamlined network design, our method achieves a good balance between performance and speed (Fig.~\ref{fig::fig1}(a)). 

To facilitate effective KD, we design three modules that operate on the input, intermediate, and output layers of the model, respectively. First, the sparse nature of 4D radar data makes it challenging for the model to extract discriminative features. To solve this, we introduce a Local Image Enhancement (LIE) module, which enhances the raw 4D radar data to produce denser representations similar to LiDAR. Next, at the intermediate layer, we propose a Feature Distribution Distillation (FDD) module, which guides the 4D radar model to learn the feature distribution of the LiDAR model. With a dual-branch design, FDD can learn from the teacher network while preserving the unique characteristics within the student model. 
At the end, a Response Distillation (RD) module is applied to the output layer, which improves the relational KD approach proposed in DistilVPR~\cite{distilvpr} by introducing a margin, further enhancing the distillation effect.

\begin{figure}[t]
        \vspace{0.15cm}
	\centering
        \includegraphics[width=1\linewidth]{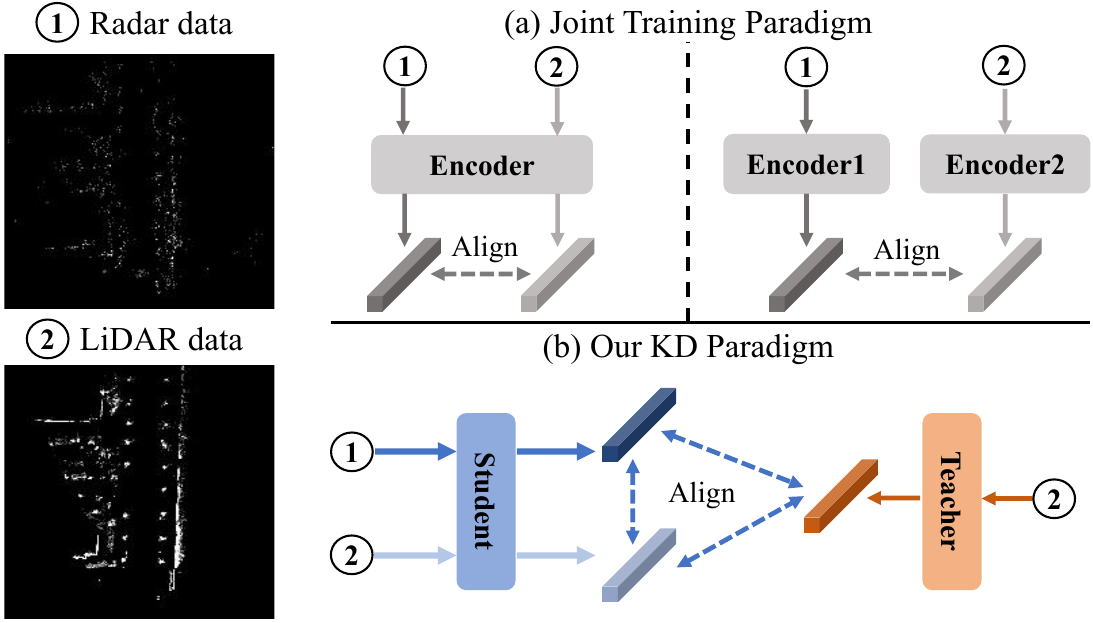}\vspace{-0.3cm}
	\caption{Comparison of the two paradigms of R2L (radar-to-LiDAR). The joint training paradigm employs one or more encoders to directly align the features of LiDAR and 4D radar, while the knowledge distillation paradigm leverages a teacher model to guide the feature alignment of two modalities.}
	\label{fig::response}
         \vspace{-0.5cm}
\end{figure}

In addition, our 4DRaL also supports 4D radar-to-LiDAR (R2L) place recognition, enabling the use of 4D radar data for place recognition within a pre-built LiDAR database. This improves flexibility in practical applications. Existing radar-to-LiDAR place recognition methods~\cite{radar-to-lidar,RaLF} typically use joint learning to align radar and LiDAR features via one or more encoders. Nevertheless, these approaches are typically inefficient since they overlook differences between the data modalities. In contrast, 4DRaL introduces KD based on joint learning to align the output features of the two modalities with those of the teacher model (Fig.~\ref{fig::response}). 
Our approach offers two advantages. First, by incorporating multiple KD modules within the model, we progressively guide the student model to generate the desired features, resulting in higher training efficiency. Second, using a high-performance teacher model and aligning its features allows our method to get superior performance. Notably, we utilize 4D radar, unlike most previous methods that rely on scanning radar. Despite limitations in the field-of-view (FOV) and resolution of 4D radar, our method still achieves excellent results.

In summary, the contributions of our work are as follows: 
\begin{itemize}
     \item We propose a novel KD framework to improve the performance of 4D radar place recognition. By leveraging the knowledge transferred from the teacher, our method achieves state-of-the-art results on benchmark datasets.
    \item To achieve effective KD, we first employ a LIE module to densify the original 4D radar data. Afterwards, a FDD module is used to guide the student to learn the feature distribution of teacher. We also propose an RD module to further enhance the efficiency of knowledge transfer.
    \item Our method can also achieve 4D radar-to-LiDAR place recognition through KD. To the best of our knowledge, it is the first work to do so in this field.
    \item We conduct extensive experiments on NTU4DRadLM
    ~\cite{ntu4dradlm} and SJTU4D~\cite{4DRadarDataset} datasets to validate the effectiveness of 4DRaL. 
    We also evaluate place recognition performance in adverse weather on the SNAIL dataset~\cite{huai2024snail}, which demonstrates the superiority of radar over LiDAR in certain extreme weather scenarios.
\end{itemize}

\section{RELATED WORK}
\subsection{Place Recognition}
\textbf{Radar Place Recognition.} 
Radar sensors mainly adopted for place recognition include scanning radar, 3D single-chip radar, and 4D radar. 
For scanning radar, several approaches have been proposed, including using handcrafted features~\cite{raplace, referee} and learned features~\cite{kidnapped, Off-the-radar}.
Although these algorithms work well on scanning radar, they may not translate directly to 4D radar due to the much lower resolution of 4D radar.
For 3D single-chip radar, AutoPlace~\cite{autoplace} extracts spatio-temporal features from radar data for robust place recognition. However, it leverages five radars simultaneously to overcome the shortcomings of sparse point clouds and limited FOV of 3D single-chip radar. Differently, mmPlace~\cite{mmplace} enhances the perception capabilities of radar by adding a rotating platform. Additionally, SPR~\cite{spr} fully utilizes radar points through extracting point-wise features, and it is also applicable to 4D radar. However, this approach inevitably increases the computational burden. For 4D radar, TransLoc4D~\cite{transloc4d} introduces the first network for 4D radar-based place recognition, exhibiting remarkable performance by leveraging the multi-modal information from radar frames. Nevertheless, the low-quality 4D radar data still limits its performance.

\textbf{Radar-to-LiDAR Place Recognition.} Radar-to-Lidar~\cite{radar-to-lidar} uses joint training to embed radar and LiDAR into a common feature space for cross-modal place recognition. Then, Yin \textit{et al}.~\cite{radar-style} leverage a generative adversarial network (GAN) to transform raw radar scans into LiDAR-like representations, enhancing localization accuracy. 
The latest work RaLF~\cite{RaLF} solves the task by using cross-modal metric learning to establish a shared embedding space between the two modalities. 
However, these algorithms are based on scanning radar, and their performance may degrade when applied to 4D radar due to its limited FOV and resolution. 
Besides, aligning the output features of modalities without considering their inherent differences may introduce irrelevant information, thereby affecting the reliability of cross-modal place recognition.
\begin{figure*}[t]
\vspace{0.15cm}
  \centering
  \includegraphics[width=1\linewidth]{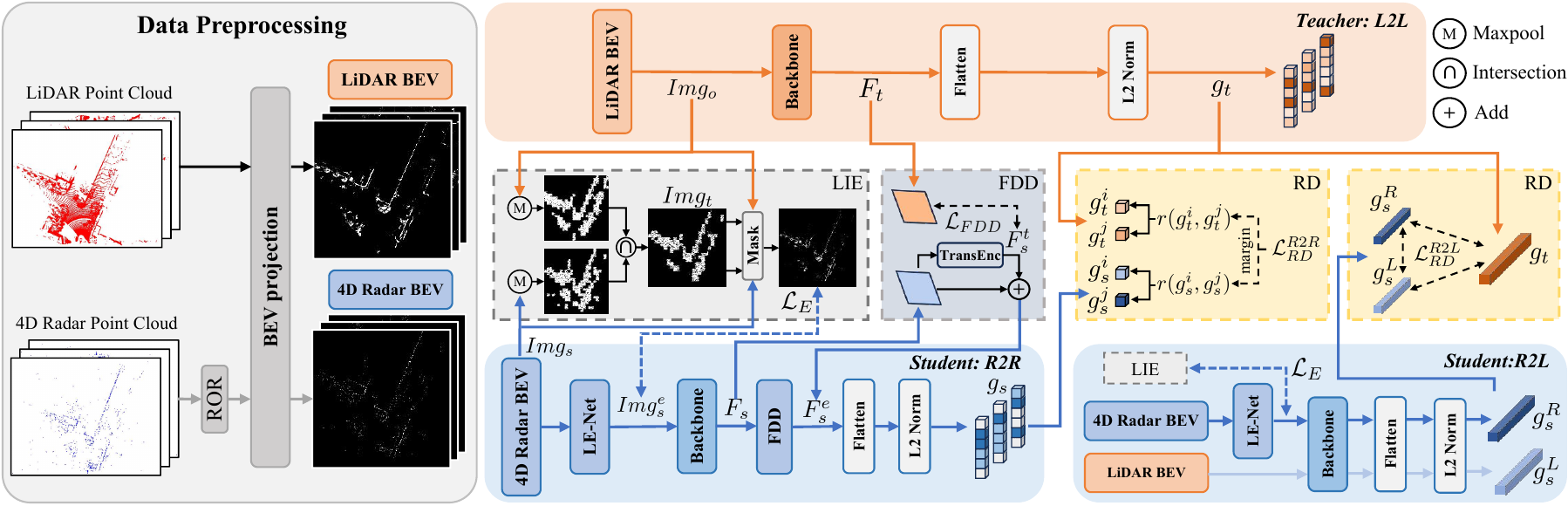}\\\vspace{-0.2cm}
  \caption{Overview of 4DRaL. Given LiDAR and 4D radar point clouds of the same scene, we generate BEV images. The teacher model extracts features from LiDAR BEV to guide the student model, which consists of R2R and R2L. R2R applies LIE for image enhancement, FDD for feature learning, and RD for refining output relationships. R2L enhances radar data with LIE and aligns LiDAR features via RD. The teacher model is used only during training.}
  \label{fig:pipeline}
  \vspace{-0.2cm}
\end{figure*}

\subsection{Knowledge Distillation}
Knowledge distillation (KD) has become a crucial technique in model compression and multi-modal learning, facilitating the transfer of knowledge from high-performance teacher networks to the small or cross-modal student models. There have been some works using KD to improve the model performance in VPR~\cite{lsdnet,distilvpr}. 
For instance, LSDNet~\cite{lsdnet} distils knowledge from a large model to a lightweight model by KD, achieving high efficiency while preserving accuracy. DistilVPR~\cite{distilvpr} employs a model that integrates LiDAR and camera as a teacher model, improving the performance of single-modal (LiDAR or camera) place recognition through KD. 
The closest method to ours is RadarDistill~\cite{radardistill}, which adopts a LiDAR-to-radar distillation path to transfer LiDAR features to radar, improving the object detection performance based on 3D single-chip radar. 
However, the architecture of RadarDistill is not suitable for place recognition since object detection focuses on detecting object-level features, whereas place recognition emphasizes the global features of the scene.

\section{METHODOLOGY}
\subsection{Data Preprocessing}
Given that 4D radar and LiDAR point clouds differ in FOV and maximum measurement distance, we first constrain their data within a common range to ensure effective KD. Next, we apply a Radius Outlier Removal (ROR)~\cite{pcl} to eliminate obvious noise in the 4D radar point cloud. Finally, we project the data of both sensors into BEV images.

\subsection{Pipeline}
We show an overview of 4DRaL in Fig.~\ref{fig:pipeline}. For each set of LiDAR and 4D radar BEV input to the model, they contain several images $\{ Img^a, Img^p,Img^n\}\in\mathbb{R}^{N\times{H}\times{W}\times{1}}$, where $N$ denotes the number of images, $H$ and $W$ mean the height and width of the image, and the final dimension represents the point cloud density in each pixel, following BVMatch [25]. The notation $(\cdot)^a$, $(\cdot)^p$, $(\cdot)^n$ represents an anchor sample, a positive sample (point clouds in the same place), and a negative sample (point clouds in a different place), respectively. We define the inputs as $Img_t$ for teacher model and $Img_s$ for student model. For R2R, $Img_t$ comprises LiDAR BEV images, while $Img_s$ consists of 4D radar BEV images. For R2L, $Img_t$ contains LiDAR BEV images, while $Img_s$ includes both 4D radar and LiDAR BEV images.

A local image enhancement (LIE) module is first applied to improve the clarity and resolution of the original 4D radar BEV image, bringing it closer to the quality of the LiDAR BEV image. The enhanced radar image $Img^e_s$ is obtained as:
\begin{equation}
Img^e_s = \text{LE-Net}(Img_s),
\label{equ:lenet}
\end{equation}
where LE-Net denotes the image enhancement network, for which we employ a lightweight U-Net~\cite{unet}. 
The features $F_t$ and $F_s$ are then extracted from the LiDAR image $Img_t$ and the enhanced radar image $Img^e_s$, respectively. We utilize ResNet18~\cite{resnet} as the backbone network, such that:
\begin{align}
F_t = \text{Backbone}(Img_t), 
F_s = \text{Backbone}(Img^e_s).
\end{align}

We design a feature distribution distillation (FDD) module to guide the student model to generate more discriminative features.
FDD takes the student feature $F_s$ as input and produces an enhanced feature representation $F^e_s$:
\begin{equation}
F^e_s = \text{FDD}(F_s).
\end{equation}

Finally, all features are flattened and normalized to obtain the global features used for place recognition, such that:
\begin{align}
g_t = \text{L2-Norm}(\text{Flatten}(F_t)) \\
g_s = \text{L2-Norm}(\text{Flatten}(F^e_s)).
\end{align}

For the global features, we also design a corresponding response distillation (RD) module, which further enhances the efficiency of knowledge transfer.

\subsection{Local Image Enhancement}  \label{sec:LIE}

A key bottleneck in 4D radar-based place recognition is its inability to supply data as rich and accurate as LiDAR. There is a method that tries to use image generation models (such as GAN) to convert radar BEV images into pseudo-LiDAR BEV images~\cite{gan}. 
However, this approach has two major limitations. First, image generation networks typically involve numerous parameters, requiring extensive data and time for training. 
Second, it fails to account for the inherent differences in perception capabilities between sensors, which limit the model's generalization ability. 
For example, radar could penetrate obstacles like smoke to detect objects behind them with its longer wavelength, whereas LiDAR cannot.

In contrast, our method emphasizes the local enhancement of images. Instead of directly using the entire LiDAR BEV for enhancement, we first generate an intermediate image that combines the 4D radar BEV and the LiDAR BEV as a mask (defined as ${Img_{o}}$).
To be specific, we apply max pooling to each image to obtain the perception areas of the two sensors, denoted as ${Img^m_s}$ and ${Img^m_t}$, which are computed as:  
\begin{align}
Img^m_s = \text{MaxPool}(Img_s), 
Img^m_t = \text{MaxPool}(Img_t).
\end{align}

The max pooling operation extracts the maximum value from each small region, effectively highlighting areas with stronger signals in the BEV image. Therefore, this approach can clearly identify the regions that contain meaningful data, while ignoring areas with sparse or zero values.

Afterwards, we determine the overlapping perception area through the intersection operation and use it as a mask ${Img_o}$: 
\begin{align}
Img_{o} &= Img^m_s \cap Img^m_t.
\end{align}

By adopting an overlapping perception area, we guarantee that the enhancement process is performed only in regions both sensors can perceive, thus addressing the differences in their perception capabilities.
After obtaining ${Img_o}$, we apply Eq. \ref{equ:lenet} to enhance $Img_s$ into $Img^e_s$, and employ a local image enhancement loss $\mathcal{L}_{E}$ to guide network training. Specifically, for each pixel in ${Img^e_s}$, if the corresponding pixel in ${Img_o}$ is non-zero, we aim to make it closer to ${Img_t}$. Otherwise, it remains unchanged, and the loss is set to zero. The local image enhancement loss $\mathcal{L}_{E}$ is formed as follows:
\begin{equation}
\small
\mathcal{L}_{E} = \frac{1}{H W} \sum_{i,j=1}^{H,W}
\mathbf{1}_{\{ Img_o(i,j) \neq 0 \}}\left( Img_t(i,j) - Img^e_s(i,j) \right)^2
\end{equation}

By using $Img_o$ as a mask, the model focuses on enhancing local areas of the image, enabling the radar BEV to produce clear contours similar to LiDAR in these regions. Additionally, preserving areas outside the overlap could significantly reduce the impact of noise during training.

\subsection{Feature Distribution Distillation}\label{sec:FDD}
Feature distillation is the process of transferring high-level feature knowledge from a teacher model to a student model. A previous method~\cite{lsdnet}
minimizes the distance between the features of teacher and student directly. This approach performs well within the same modality, since there are typically no significant modality differences between the teacher and student models. Although 4D radar and LiDAR data are both represented as point clouds, they show obvious differences in resolution, perception range, etc. To achieve effective feature-level distillation, we propose feature distribution distillation.

We propose a dual-branch structure with inputs as $F_s$. One branch learns the high-level features of the teacher model, while the other branch preserves the original features of the student model. With this design, the student model can both retain its unique feature representation and learn the features of the teacher model, thereby avoiding complete imitation.

Specifically, we introduce a feature transformation module named TransEnc to align the feature distribution of $F_s$ with $F_t$. Thus, the transformed features $F^t_s$ are obtained as:
\begin{equation}
F^t_s = \text{TransEnc}(F_s).
\end{equation}

TransEnc first applies a 2D convolutional layer with a rectified linear unit (ReLU) to process $F_s$, followed by another 2D convolution layer for further feature transformation.
Next, we utilize Kullback-Leibler (KL) divergence to calculate the loss for this branch:  
\begin{equation}
\mathcal{L}_{FDD} = D_{\text{KL}}\left( F_t \ || \ F^t_s \right),
\end{equation}
where $D_{\text{KL}}$ represents the KL divergence. 
The other branch maintains the original features $F_s$ of the student model unchanged, which are combined with the transformed features $F^t_s$ to obtain the enhanced feature $F^e_s$:
\begin{equation}
    F^e_s = F_s + F^t_s.
\end{equation}

\subsection{Response Distillation}\label{sec:RD}
Response distillation transfers knowledge by aligning final outputs of teacher and student models. We refer to the relational KD method in DistilVPR~\cite{distilvpr}, which transfers implicit knowledge by measuring the relationships between samples:
\begin{equation}\mathcal{L}_{\text{DistilVPR}} = \ell(r(g^i_t, g^j_t), r(g^i_s, g^j_s)),
\end{equation}
where $\ell(\cdot)$ denotes the loss function (e.g., L1, L2), and $r(\cdot,\cdot)$ represents the relational function that computes the distance between features (e.g., Euclidean distance). $(\cdot)^i$ and $(\cdot)^j$ refer to two different samples.

Nevertheless, the original relational KD method does not work well with our model. As displayed in Fig.~\ref{fig::output}, we observe that the original relational KD improves convergence speed and performance in the early stages of training compared to not using it. 
However, the performance starts to decline as the number of training iterations increases. 
This suggests that the student model struggles to fully replicate the relationships between the output samples of the teacher model because of inherent differences in the data distributions.

To address this problem, we introduce a margin into the original relational KD method:
\begin{equation}
    \mathcal{L}^{R2R}_{RD} = \text{Max}\{ \mathcal{L}_{\text{DistilVPR}} - m_{RD}^{R2R}, 0\},
\end{equation}
where $m_{RD}^{R2R}$ represent the margin. In $\mathcal{L}_{\text{DistilVPR}}$, we adopt the Mean Squared Error (MSE) for $\ell(\cdot)$ and the Euclidean distance for $r(\cdot,\cdot)$. 

By adding a margin, we mitigate the adverse effects caused by these distributional differences. The margin provides a buffer that prevents the student model from overly and rigidly mimicking the relationships in the teacher model, allowing it to better generalize to its own data distribution. This adjustment encourages the student model to learn more flexible yet robust relational patterns, improving overall training stability and performance.

\subsection{4D radar-to-LiDAR Place Recognition}
For R2R, we have used KD to guide 4D radar in generating LiDAR-like features and improved the network performance. Building on this, we further explore using these LiDAR-like features from 4D radar to achieve R2L place recognition.

The goal of R2L is to align the features of 4D radar and LiDAR at the same location for accurate place recognition. Since it emphasizes the similarity of features between the two modalities, we modify certain module configurations in R2R. For R2L, LIE (as described in Section \ref{sec:LIE}) is also introduced since it can significantly improve the density of 4D radar BEV images. FDD (as described in Section \ref{sec:FDD}) is not used because it largely preserves features within its own modality, which is detrimental to the R2L task. To strengthen the similarity between the 4D radar and LiDAR features generated by the student model, we adjust the RD module in R2R. Specifically, we guide the output features of 4D radar and LiDAR in the student model to align with the features of teacher. Consistent with previous methods~\cite{radar-to-lidar,RaLF}, we also align the 4D radar features and LiDAR features output by the student model itself. Similarly, we introduce margins to ensure the stable training of the model:
\begin{align}
    \mathcal{L}^{R2L}_{RD} = &\ \text{Max}\{\|g^R_s - g_t\|_2 - m_{RD}^{R2L}, 0\} \nonumber \\
    &+ \text{Max}\{\|g^L_s - g_t\|_2 - m_{RD}^{R2L}, 0\} \\
    &+ \text{Max}\{\|g^L_s - g^R_s\|_2 - m_{RD}^{R2L}, 0\} \nonumber,
\end{align}
where $g^R_s$ and $g^L_s$ represent the global features of 4D radar and LiDAR output by the student model, respectively.

\subsection{Loss Functions}
The total loss function used to train R2R is defined as:
\begin{equation}
    \mathcal{L}^{R2R}_{total} = \mathcal{L}^{R2R}_{triplet} + \mathcal{L}_E + \mathcal{L}_{FDD} + \mathcal{L}_{RD}^{R2R},
\end{equation}
where $\mathcal{L}^{R2R}_{triplet}$ is the triplet loss~\cite{schroff2015facenet}, which is generally used in place recognition tasks.
\begin{equation}
    \mathcal{L}^{R2R}_{triplet} = \text{Max}\{d(g^a_s,g^p_s)-d(g^a_s,g^n_s) + m^{R2R}_{triplet}, 0 \}
\end{equation}
where $m^{R2R}_{triplet}$ represents the desired separation margin.

The total loss function applied to train R2L is as follows:
\begin{equation}
    \mathcal{L}^{R2L}_{total} = \mathcal{L}^{R2L}_{triplet} + \mathcal{L}_E + \mathcal{L}_{RD}^{R2L},
\end{equation}
where the $\mathcal{L}^{R2L}_{triplet}$ used for R2L is obtained from:
\begin{equation}
    \mathcal{L}^{R2R}_{triplet} = \text{Max}\{d(g^a_s,g^p_t)-d(g^a_s,g^n_s) + m^{R2L}_{triplet}, 0 \}.
\end{equation}

We substitute $g^p_s$ with $g^p_t$, allowing the output features of the student model to more effectively align with those of the teacher model.

\section{EXPERIMENTS}
\subsection{Experimental Settings}
\textbf{Benchmark Datasets.} 
We adopt the benchmark datasets proposed by TransLoc4D~\cite{transloc4d} for a fair comparison. The benchmark datasets are designed for the 4D radar place recognition task. It includes two datasets: the NTU4DRadLM dataset (NTU4D)~\cite{ntu4dradlm} and the SJTU4D dataset~\cite{4DRadarDataset}. NTU4D employs an Oculii Eagle 4D radar and a Livox Horizon 3D LiDAR, covering approximately 17.6 kilometers of data. SJTU4D utilizes a ZF FRGen21 4D radar and a Hesai Panda128 LiDAR, with data collected from various environments, such as an industrial zone and a university campus. The benchmark dataset also includes three new subsequences collected from the same place as the NTU4D dataset. However, these subsequences only provide normalized 4D radar point clouds. Due to the absence of scale information, we are unable to convert the data into BEV images. The details of the benchmark datasets are shown in Table~\ref{tab:allocation}. 

\begin{table}[t]
\centering
\LARGE
\caption{The details of the benchmark datasets.}
\vspace{-0.1cm}
\resizebox{1\linewidth}{!}{
\begin{tabular}{c|cc|c|cc}
\toprule[.11cm]
Dataset & Splits & Frame Num. &Dataset & Splits & Frame Num. \\
\hline \rule{0pt}{8pt}
\multirow{4}{*}{NTU4D} & train\_query & 7620 & \multirow{4}{*}{SJTU4D} & test\_a\_query & 7634  \\
~ & train\_database & 10000 & ~ & test\_a\_database & 7500  \\
~ & validation\_query & 7002 & ~ & test\_b\_query & 6501  \\
~ & validation\_database & 10839 & ~ & test\_b\_database & 2500 \\
\toprule[.11cm]
\end{tabular}
}
\label{tab:allocation}
\vspace{-0.6cm}
\end{table}

\begin{table*}[!t]
\small
\renewcommand\tabcolsep{10.0pt}
\vspace{-0.06in}
\caption{Comparison with state-of-the-art of R2R in benchmark datasets.}
\vspace{-0.075in}
~\label{tab:Sipailou Campus}
\centering
\resizebox{\linewidth}{!}{
\begin{tabular}{c|c|ccc|ccc|ccc|cc}
\toprule[.06cm]
\multirow{2}{*}{Methods}  & \multirow{2}{*}{Modality} & \multicolumn{3}{c|}{NTU4D}& \multicolumn{3}{c|}{SJTU4D-TestA} & \multicolumn{3}{c|}{SJTU4D-TestB} & Runtime & Param\\ \cline{3-5} \cline{6-8} \cline{9-11} \rule{0pt}{8pt}

~ & & R@1 & R@5 & R@10 & R@1 & R@5 & R@10 & R@1 & R@5 & R@10 & ms & \(\cdot 10^6\)\\ 
\hline 
\rule{0pt}{8pt}
4DRaL (Teacher) & L2L & 97.2 & 97.5 & 98.0 & 93.8 & 94.4 & 94.5 & 89.0 & 90.2 & 92.4 & 1.2 &2.78\\ \hline \rule{0pt}{8pt}
In. Scan Context~\cite{wang2020intensity} & R2R  & 84.1 & 86.2 & 90.2 & 67.9 & 79.8 & 83.7 & 78.0 & 86.5 & 90.2 & 0.2 & \(10^{-5}\)\\ 
TransLoc3D~\cite{transloc3d} & R2R & 87.0 & 90.2 & 92.6 & 75.6 & 89.2 & 92.2 & 79.3 & 86.0 & 87.5 & 11.2 &6.63\\
MinkLoc3Dv2~\cite{minkloc3dv2} & R2R & 91.5  & 93.8 & 95.2 & 88.6 & 93.2 & \textbf{94.0} & 85.8 & 87.9 & 88.9 & 8.8 & 2.66\\ 
PTC-Net-L~\cite{ptcnet} & R2R & 95.1  & 95.9 & \textbf{96.5} & 79.8 & 89.6 & 91.4 & 80.3 & 85.3 & 86.8 & 4.2 & 0.08 \\ 
Autoplace~\cite{autoplace} & R2R & 94.5 & 95.2 & 96.1 & 80.5 & 88.8 & 91.2 & 80.2 & 84.9 & 86.5 & 0.6 & 2.34\\ 
RaLF~\cite{RaLF} & R2R & 95.2 & 95.6 & 95.8 & 89.1 & 92.5 & 93.5 & 84.2 & 87.0 & 87.2 & 2.6 & 2.97\\ 
TransLoc4D~\cite{transloc4d} & R2R & 94.8 & 95.9 & 96.3 & 90.8 & 92.9 & 93.4 & 85.9 & 88.7 & \textbf{90.5} & 12.1 & 3.35\\ \hline  \rule{0pt}{8pt}
4DRaL (w/o KD) & R2R & 94.5 & 95.1 & 95.5 & 86.7 & 90.2 & 91.5 & 84.4 & 86.8 & 87.7 & 1.2 & 2.78\\ 
4DRaL (w/ KD) & R2R & \textbf{96.1} & \textbf{96.3} & \textbf{96.5} & \textbf{91.6} & \textbf{93.4} & \textbf{94.0} & \textbf{86.3} & \textbf{89.8} & 90.3  & 2.6 & 3.86\\ 
\textit{Improvement} & - & \textcolor[rgb]{0.0,0.5,0.0}{\textit{$\uparrow$1.6}} & \textcolor[rgb]{0.0,0.5,0.0}{\textit{$\uparrow$1.2}}  & \textcolor[rgb]{0.0,0.5,0.0}{\textit{$\uparrow$1.0}} & \textcolor[rgb]{0.0,0.5,0.0}{\textit{$\uparrow$4.9}} & \textcolor[rgb]{0.0,0.5,0.0}{\textit{$\uparrow$3.2}} & \textcolor[rgb]{0.0,0.5,0.0}{\textit{$\uparrow$2.5}} & \textcolor[rgb]{0.0,0.5,0.0}{\textit{$\uparrow$1.9}} & \textcolor[rgb]{0.0,0.5,0.0}{\textit{$\uparrow$3.0}} & \textcolor[rgb]{0.0,0.5,0.0}{\textit{$\uparrow$2.6}} & - & - \\
\toprule[.06cm]
\end{tabular}
}
\normalsize
\label{tab:result}
\vspace{-0.1cm}
\end{table*}

\begin{table*}[!t]
\small
\renewcommand\tabcolsep{10.0pt}
\vspace{-0.06in}
\caption{Comparison with state-of-the-art of R2L in benchmark datasets.}
\vspace{-0.075in}
~\label{tab:Sipailou Campus}
\centering
\resizebox{\linewidth}{!}{
\begin{tabular}{c|c|ccc|ccc|ccc|cc}
\toprule[.06cm]
\multirow{2}{*}{Methods}  & \multirow{2}{*}{Modality} & \multicolumn{3}{c|}{NTU4D}& \multicolumn{3}{c|}{SJTU4D-TestA} & \multicolumn{3}{c|}{SJTU4D-TestB} & Runtime & Param\\ \cline{3-5} \cline{6-8} \cline{9-11} \rule{0pt}{8pt}

~ & & R@1 & R@5 & R@10 & R@1 & R@5 & R@10 & R@1 & R@5 & R@10 & ms & \(\cdot 10^6\)\\ 
\hline 
\rule{0pt}{8pt}
4DRaL (Teacher) & L2L & 97.2 & 97.5 & 98.0 & 93.8 & 94.4 & 94.5 & 89.0 & 90.2 & 92.4 & 1.2 &2.78\\ \hline \rule{0pt}{8pt}
Radar-to-LiDAR~\cite{radar-to-lidar} & R2L & 14.3 & 21.0 & 27.4  
  & 4.0 & 7.1 & 13.0  
  & 19.8 & 36.3 & 39.6 & 2.0 & 4.28\\
RaLF~\cite{RaLF} & R2L & 22.4 & 27.5 & 34.3  
  & 27.4 & 40.8 & 46.8  
  & 49.6 & 64.8 & 68.1 & 2.6 & 2.97\\ \hline  \rule{0pt}{8pt}
4DRaL (w/o KD) & R2L & 16.5 & 22.5 & 28.9 
  & 13.3 & 27.2 & 36.3  
  & 32.3 & 46.1 & 53.5 & 1.2 & 2.78\\ 
4DRaL (w/ KD) & R2L & \textbf{34.8} & \textbf{43.5} & \textbf{48.7}  
  & \textbf{38.2} & \textbf{49.8} & \textbf{58.1}  
  & \textbf{75.5} & \textbf{80.3} & \textbf{83.2} & 2.5 & 3.46\\ 
\textit{Improvement} & - & \textcolor[rgb]{0.0,0.5,0.0}{\textit{$\uparrow$18.3}} & \textcolor[rgb]{0.0,0.5,0.0}{\textit{$\uparrow$21.0}}  & \textcolor[rgb]{0.0,0.5,0.0}{\textit{$\uparrow$19.8}} & \textcolor[rgb]{0.0,0.5,0.0}{\textit{$\uparrow$24.9}} & \textcolor[rgb]{0.0,0.5,0.0}{\textit{$\uparrow$22.6}} & \textcolor[rgb]{0.0,0.5,0.0}{\textit{$\uparrow$21.8}} & \textcolor[rgb]{0.0,0.5,0.0}{\textit{$\uparrow$43.2}} & \textcolor[rgb]{0.0,0.5,0.0}{\textit{$\uparrow$34.2}} & \textcolor[rgb]{0.0,0.5,0.0}{\textit{$\uparrow$29.7}} & - & -\\ 

\toprule[.06cm]
\end{tabular}
}
\normalsize
\label{tab:r2l_result}
\vspace{-0.5cm}
\end{table*}

\begin{table}[t]
\large
\renewcommand\tabcolsep{10.0pt}
\caption{Ablation studies to evaluate the KD module composition.}
\vspace{-0.1in}
\centering
\resizebox{\linewidth}{!}{
\begin{tabular}{c|c|ccc|ccc}
\toprule[.08cm]
\multirow{2}{*}{Methods}  & \multirow{2}{*}{Modality} & \multirow{2}{*}{$\mathcal{L}_{E}$} & \multirow{2}{*}{$\mathcal{L}_{FDD}$} & \multirow{2}{*}{$\mathcal{L}_{RD}$} &  \multicolumn{3}{c}{NTU4D} \\ \cline{6-8}  \rule{0pt}{8pt}

~ & ~ & ~ & ~ & ~ & R@1 & R@5 & R@10\\ \hline \rule{0pt}{8pt}

\hspace{-3pt}a0 & R2R & ~ & ~ & ~ & 94.5 & 95.1 & 95.5 \\ 
a1 & R2R & $\checkmark$ & ~ & ~ &95.2 & 95.6 &96.0\\ 
a2 & R2R & $\checkmark$ & $\checkmark$ & ~ &95.7 &96.1 &96.3\\ 
a3 & R2R & $\checkmark$ & $\checkmark$ & $\checkmark$ & \textbf{96.1} & \textbf{96.3} & \textbf{96.5} \\  \hline \rule{0pt}{8pt}
\hspace{-3pt}b0 & R2L & ~ & ~ & ~ & 16.5 & 22.5 & 28.9 \\ 
b1 & R2L & $\checkmark$ & ~ & ~ &28.4 & 35.2 & 44.6 \\
b2 & R2L & $\checkmark$ & $\checkmark$ & ~ &22.5 &28.2 &38.6 \\
b3 & R2L & $\checkmark$ & ~ & $\checkmark$ & \textbf{34.8} & \textbf{43.5} & \textbf{48.7}\\ 			
\toprule[.08cm]
\end{tabular}
}
\label{tab:kd_composition} 
\vspace{-0.4cm}
\end{table}

\textbf{Evaluation Metric.} Following previous methods~\cite{autoplace,transloc4d}, we adopt the $Recall\text{@}N (N\in\{1,5,10\})$ metric to evaluate the performance of different place recognition algorithms. Referring to TransLoc4D~\cite{transloc4d}, we regard a retrieval 4D radar point cloud as correct if at least one of the top $N$ retrieved candidates is within 25 meters of the Euclidean distance.

\textbf{Implementation Details.} For the 4D radar and LiDAR point cloud, we retain points within the range of [0m, 80m] on the X-axis and [-40m, 40m] on the Y-axis, transforming them into BEV images with the resolution of 200 × 200. The max pooling operation (Section \ref{sec:LIE}) uses a 3×3 kernel with stride 1 and padding 1. The TransEnc module (Section \ref{sec:FDD}) includes two 2D convolutional layers. The first takes 256 input channels and outputs 128, while the second takes 128 and outputs 256. Both layers adopt a 3×3 kernel, stride 1, and padding 1.
The distillation and triplet margins are uniformly set as $m_{RD}^{R2R} = m_{RD}^{R2L} = 0.01$ and $m_{triplet}^{R2R} = m_{triplet}^{R2L} = 0.3$. In the training phase, positive samples are selected within a 10m radius, while negative samples are chosen from outside a 50m radius. We choose 1 positive sample and 4 negative samples to calculate the triplet loss. Both R2R and R2L models are trained for 40 epochs on an NVIDIA RTX 4090 GPU, utilizing the Adam optimizer with a learning rate of 1 $\times$ 10$^{-3}$. 

\subsection{Comparison with the State-of-the-art Methods} 
To guarantee a fair comparison, all learning-based methods are trained on the training set of benchmark datasets and use only a single-frame point cloud.

\textbf{R2R Place Recognition.} Since place recognition methods specifically designed for 4D radar are still limited, we further validate the effectiveness of 4DRaL by comparing it with methods based on LiDAR, 3D single-chip radar, and scanning radar. The comparative methods include Intensity Scan Context~\cite{wang2020intensity}, TransLoc3D~\cite{transloc3d}, MinkLoc3Dv2~\cite{minkloc3dv2}, PTC-Net~\cite{ptcnet}, Autoplace~\cite{autoplace}, RaLF~\cite{RaLF}, TransLoc4D~\cite{transloc4d}. Table~\ref{tab:result} shows the experimental results of all methods. The sparsity of radar points makes it challenging for handcrafted descriptors to extract discriminative features, leading to limited performance of Intensity Scan Context~\cite{wang2020intensity}. 
Among LiDAR-based place recognition approaches~\cite{transloc3d, minkloc3dv2, ptcnet}, although they generally perform well in individual sequences, their overall performance remains relatively average. Autoplace~\cite{autoplace} relies on multiple data frames and performs poorly when limited to a single frame. 
RaLF~\cite{RaLF} surpasses Autoplace by using joint training with LiDAR data. TransLoc4D~\cite{transloc4d} exhibits remarkable performance but remains constrained by the low-quality radar points. 
In contrast, 4DRaL exceeds TransLoc4D on most metrics using a high-performance teacher model for guidance.
Table~\ref{tab:result} also shows the improvement in model performance using KD. Although the improvements on NTU4D are all around 1\%, the highest performance improvement on the SJTU4D reaches 4.9\% (SJTU4D-TestA R@1). 
This is because the radar points in the SJTU4D dataset are sparser than those in the NTU4D, which amplifies the effect of KD.

\textbf{R2L Place Recognition.} Since no methods are dedicated to this task, we compare 4DRaL with other methods based on scanning radar, all of which are retrained using benchmark datasets. We adopt Radar-to-LiDAR~\cite{radar-to-lidar} and RaLF~\cite{RaLF} for comparison. Table~\ref{tab:r2l_result} indicates the experimental results. The results of Radar-to-LiDAR~\cite{radar-to-lidar} demonstrate that simple joint training does not obtain good results, and RaLF gets better performance with multiple encoders. 
In contrast, our method achieves the best performance across all sequences through KD to guide the training of model. 
Similar to the R2R task, the performance improvement is more pronounced on the SJTU4D, particularly for the R@1 metric on the SJTU4DPR-TestB, where a significant 43.2\% increase is achieved by KD.

\begin{table}[t!]
\centering
\renewcommand{\arraystretch}{0.9}
\caption{Ablation studies on Local Image Enhancement.}
\vspace{-0.22cm}
\label{tab:local}
\scriptsize 
\setlength{\tabcolsep}{14pt} 
\begin{tabular}{c|c|ccc}
\toprule[.05cm]
\multirow{2}{*}{Methods} & \multirow{2}{*}{Modality} & \multicolumn{3}{c}{NTU4D}  \\ \cline{3-5} \rule{0pt}{8pt}
~ & & R@1 & R@5 & R@10 \\ 
\hline 
\rule{0pt}{8pt}
Raw    & R2R & 94.5 & 95.1 & 95.5\\ 
Global & R2R & 94.9 & 95.6 & 96.2 \\ 
Local  & R2R & \textbf{96.1} & \textbf{96.3} & \textbf{96.5}\\ \hline \rule{0pt}{8pt}
Raw    & R2L & 16.5 & 22.5 & 28.9\\ 
Global  & R2L & 25.7 & 30.9 & 38.2 \\ 
Local  & R2L & \textbf{34.8} & \textbf{43.5} & \textbf{48.7}\\ 
\toprule[.05cm]
\end{tabular}
\vspace{-0.7cm}
\end{table}

\textbf{Efficiency Analysis.} Table~\ref{tab:result} and Table~\ref{tab:r2l_result} also illustrate the number of parameters and runtime for all methods. For R2R, 4DRaL is four times faster than the previous best work (TransLoc4D~\cite{transloc4d}), requiring only 2.6 milliseconds (ms) per run. We think that the efficiency should be attributed to two factors: the use of the BEV point cloud representation and the simplicity of our convolutional module. For R2L, 4DRaL maintains nearly the same speed as RaLF [16] despite a slight increase in parameters.
This demonstrates a good balance between performance and speed of 4DRaL.

\subsection{Ablation Studies}
In this section, we also conduct a series of ablation studies to thoroughly explore the effects of the different KD module components and their configurations in our proposed 4DRaL.

\textbf{KD Module Composition.} Our algorithm includes three modules: LIE, FDD, and RD. Each module corresponds to a specific loss: $\mathcal{L}_E$, $\mathcal{L}_{FDD}$, and $\mathcal{L}_{RD}$. Table~\ref{tab:kd_composition} illustrates the influence of various KD module combinations on model performance. LIE (a1, b1) improves performance in both R2R and R2L compared to the original approach (a0, b0). This improvement is attributed to the enhanced 4D radar BEV images attaining clarity comparable to LiDAR. Then, FDD is further introduced. In R2R, FDD enables the model to generate more discriminative features and further enhances model performance (a2). However, in R2L, the performance is severely degraded (b2) as FDD retains an excessive amount of intra-modality features, reducing the similarity between inter-modality features. Finally, RD aligns the output feature of the student model with that of the teacher model, enhancing the generalization ability in both R2R (a3) and R2L (b3). 

\begin{table}[t!]
\centering
\renewcommand{\arraystretch}{1.2} 
\caption{Ablation studies on Feature Distribution Distillation.}
\vspace{-0.2cm}
\label{tab:FDD}
\scriptsize
\setlength{\tabcolsep}{14pt}
\begin{tabular}{cc|ccc}
\toprule[.05cm]
\multicolumn{2}{c|}{\multirow{2}{*}{Method}} & \multicolumn{3}{c}{NTU4D} \\ \cline{3-5} 
\rule{0pt}{10pt} & & R@1 & R@5 & R@10 \\ \hline 
\multicolumn{2}{c|}{\rule{0pt}{10pt}Raw} & 94.5 & 95.1 & 95.5 \\ \hline
\rule{0pt}{10pt} \multirow{2}{*}{Single} & KL & 95.3 & 95.9 & 96.2 \\ 
 & MSE & 95.0 & 95.5 & 96.2 \\ \hline 
\rule{0pt}{10pt} \multirow{2}{*}{Dual} & KL & \textbf{96.1} & \textbf{96.3} & \textbf{96.5} \\ 
 & MSE & 95.8 & 96.1 & 96.4 \\ \bottomrule[.05cm]
\end{tabular}
\vspace{-0.45cm}
\end{table}
\begin{figure}[t!]
\centering
\includegraphics[width=0.9\linewidth]{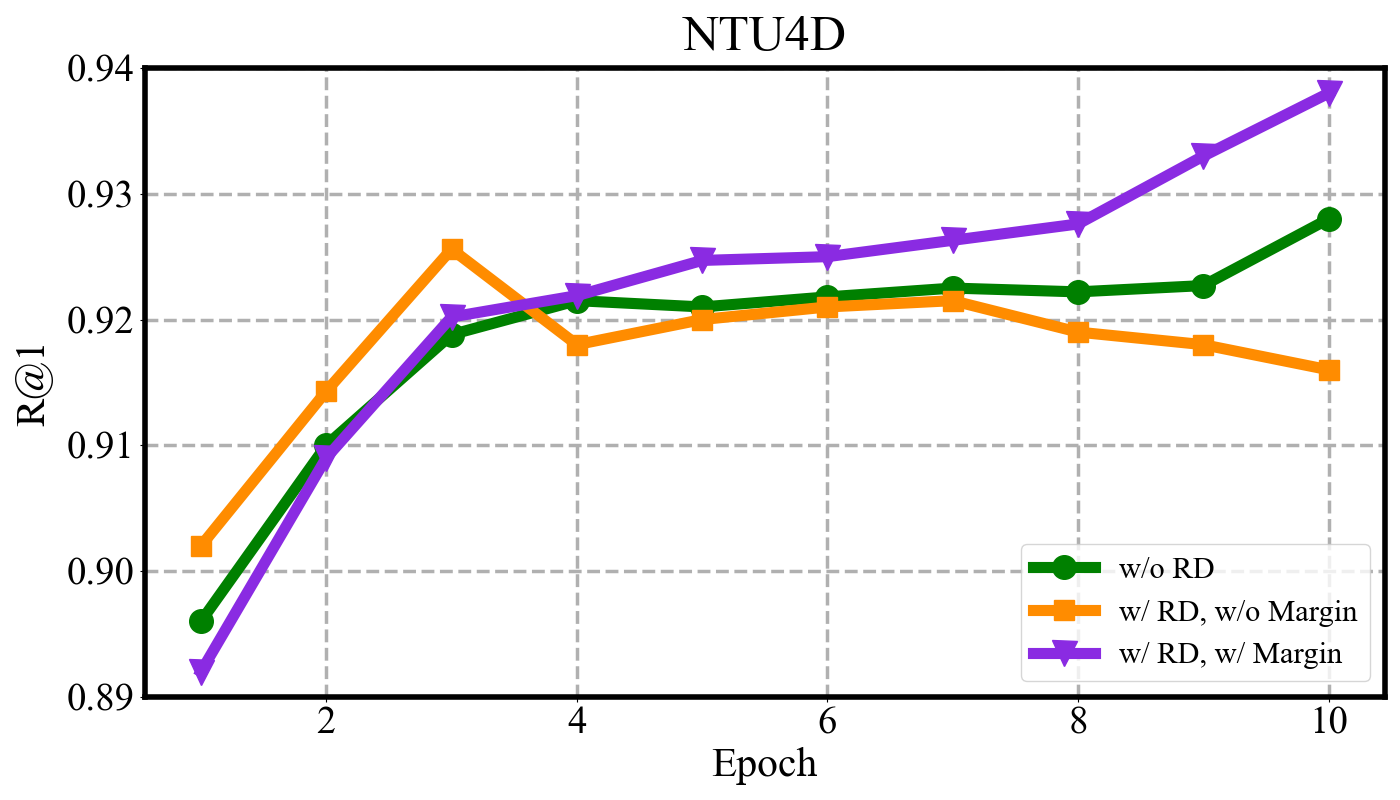}
\vspace{-0.15in}
\caption{The influence of margin on Response Distillation (RD) module.}
\label{fig::output}
\vspace{-0.27cm}
\end{figure}

\begin{figure}[t!]
\centering
\includegraphics[width=0.95
\linewidth]{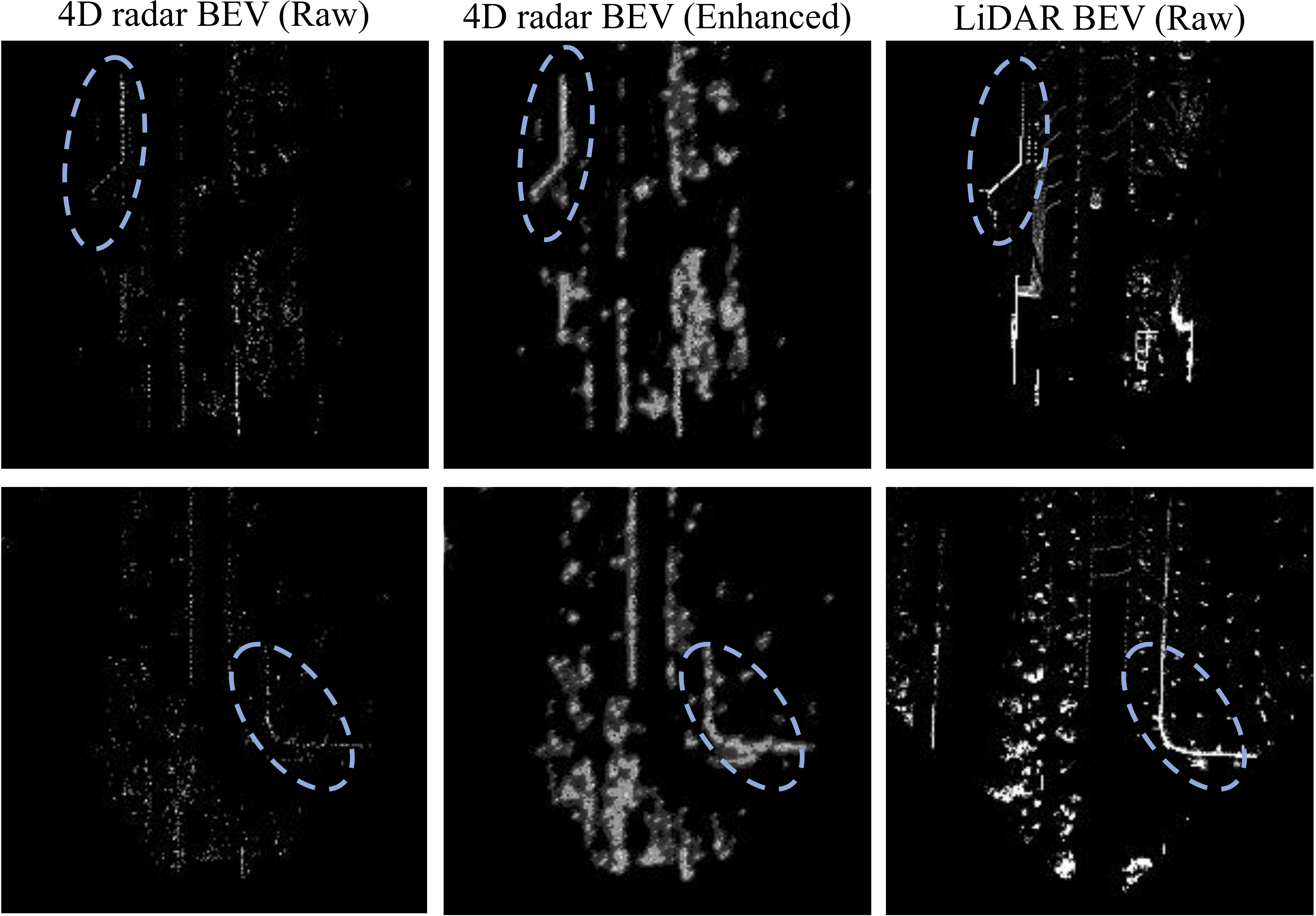}\vspace{-0.3cm}
\caption{Visualization of Local Image Enhancement. In the area highlighted by the blue ellipse, the enhanced 4D Radar BEV image presents a clearer outline compared to the original image.}
\label{fig::finnal}
\vspace{-0.37cm}
\end{figure}

\textbf{Local Image Enhancement.} Previous methods often use the entire LiDAR BEV to enhance the radar BEV image~\cite{gan}. To further investigate the impact of enhancement method, we conduct ablation studies to explore the effects of both global and local enhancement. In Table~\ref{tab:local}, the global enhancement performs worse than the local enhancement in both R2R and R2L place recognition tasks. This proves that exploiting local cues is more beneficial than global ones for radar-LiDAR cross-modal enhancement.

\begin{figure}[t]
\centering
\includegraphics[width=0.95
\linewidth]{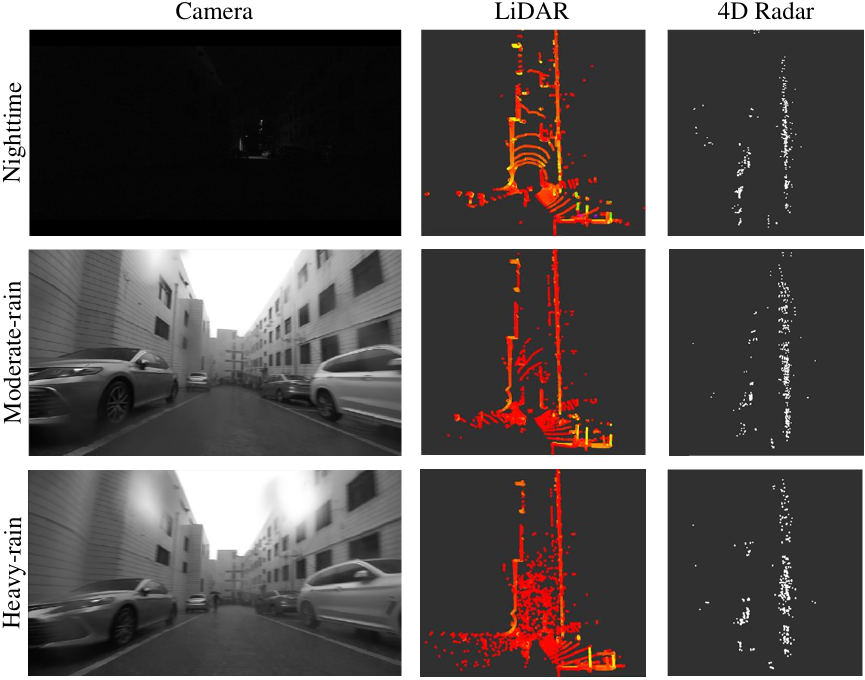}
\vspace{-0.3cm}
\caption{Visualization of LiDAR and 4D radar data across different weather conditions in the SNAIL radar dataset.}
\label{fig::rain}
\vspace{-0.05cm}
\end{figure}

\textbf{Feature Distribution Distillation.} We introduce a dual-branch structure in FDD. To validate its effectiveness, we compare the dual-branch and single-branch structures. Additionally, we also conduct experiments by replacing the KL loss with MSE (Mean Squared Error) loss. As displayed in Table~\ref{tab:FDD}, the single-branch structure does not yield substantial improvement, indicating that directly mapping the output features of the student model to that of the teacher model is inefficient. In contrast, the dual-branch structure significantly enhances the performance, and the KL loss also demonstrates better generalization than MSE loss. This occurs because KL loss focuses on optimizing the distribution difference rather than merely minimizing the point-to-point error.

\textbf{Response Distillation.} For 4DRaL, we employ a margin based on the relational distillation of DistilVPR~\cite{distilvpr}. In order to intuitively display the impact of margin, we visualize the model performance change curve of the first 10 epochs in the training process on the NTU4D sequence (as shown in Fig.~\ref{fig::output}). Without the margin, the R@1 metric initially rises, then declines, and eventually stabilizes, while after adding margin, the performance steadily and continuously improves. 
This suggests that over-reliance on teacher may not achieve optimal results, and introducing a margin assists the student in retaining its unique traits, thereby enhancing performance. 

\textbf{Visualization of Local Image Enhancement.}
To exhibit the influence of the LIE module on 4D radar BEV images, we present images before and after enhancement. As shown in Fig.~\ref{fig::finnal}, the enhanced radar images display clearer details and achieve a resolution comparable to the LiDAR BEV images, particularly in the area highlighted by the blue ellipse. This improvement validates the effectiveness of our LIE approach. By refining regional details and resolution, the enhancement facilitates the extraction of discriminative features, ultimately boosting the performance of place recognition.

\subsection{Evaluation Under Different Weather Conditions}

\begin{table}[t!]
\renewcommand\tabcolsep{1.5pt}
\scriptsize
\caption{Comparison of R2R and R2L performance on the SNAIL.}
\vspace{-0.1in}
~\label{tab:Sipailou Campus}
\centering
\begin{tabular}{c|c|ccc|ccc|ccc}
\toprule[.05cm]
\multirow{2}{*}{Methods}  & \multirow{2}{*}{Mod.} & \multicolumn{3}{c|}{SS-night}& \multicolumn{3}{c|}{SS-moderate-rain} & \multicolumn{3}{c}{SS-heavy-rain}\\ \cline{3-5} \cline{6-8} \cline{9-11} \rule{0pt}{8pt}

~ & & R@1 & R@5 & R@10 & R@1 & R@5 & R@10 & R@1 & R@5 & R@10\\ 
\hline 
\rule{0pt}{8pt}
RaLF & R2R & 82.9 & 90.5 & 93.8 & 84.8 & 91.5 & 95.7 & 81.2 & 89.6 & 94.3 \\ 
TransLoc4D & R2R & 84.5 & 88.1 & 91.5 & 81.9 & 91.1 & 93.3 & 79.8 & 84.3 & 90.4 \\
4DRaL (Ours) & R2R & \textbf{88.4} & \textbf{93.3} & \textbf{95.9} & \textbf{90.1} & \textbf{95.7} & \textbf{97.4} & \textbf{88.1} & \textbf{93.5} & \textbf{96.0} \\ \hline \rule{0pt}{8pt}
Radar-to-LiDAR & R2L & 17.8 & 30.8 & 38.1 & 16.2 & 28.4 & 33.4 & 20.8 & 34.8 & 45.6\\
RaLF & R2L & 25.2 & 36.7 & 42.8 & 28.9 & 42.8 & 57.2 & 30.5 & 44.8 & 60.4 \\ 
4DRaL (Ours) & R2L & \textbf{36.5} & \textbf{46.8} & \textbf{59.3} & \textbf{40.2} & \textbf{54.3} & \textbf{62.1} & \textbf{37.5} & \textbf{54.2} & \textbf{62.5} \\ 

\toprule[.05cm]
\end{tabular}
\label{tab:result_r2l}
\vspace{-0.7cm}
\end{table}

To further demonstrate 4DRaL's robustness in adverse weather, we conduct additional evaluations utilizing the software school (SS) sequence of SNAIL dataset~\cite{huai2024snail}, with the daytime sequence (20231109/4) to build the database, and the nighttime (20231019/2) and rainy sequences (20231105/5, 20231105/4) as the query set. Moreover, to evaluate 4DRaL's generalization ability, we use a model pre-trained on the TransLoc4D benchmark~\cite{transloc4d} for place recognition, without performing any fine-tuning.


As shown in Table~\ref{tab:result_r2l}, 4D radar demonstrates consistent robustness to environmental variations in both R2R and R2L tasks. Notably, TransLoc4D's performance degrades on the SNAIL dataset due to its velocity-dependent features, which prove inconsistent under diverse weather. In contrast, our knowledge distillation framework extracts more robust features, yielding superior performance and generalization.

To validate the practical advantage of R2R place recognition in challenging environments, we conduct an extra L2L experiment based on our algorithm in Table~\ref{tab:lidarvsradar}. Although LiDAR exceeds radar under normal weather, its performance declines in rain. In moderate and heavy rain, LiDAR performance degrades significantly due to noise, whereas 4D radar remains stable and even outperforms LiDAR, demonstrating its superior robustness under adverse weather conditions.

\begin{table}[!t]
\centering
\scriptsize 
\setlength{\tabcolsep}{3.5pt} 
\caption{Comparison of LiDAR-to-LiDAR and 4D radar-to-4D radar place recognition performance under adverse weather.}
\vspace{-0.25cm}
\label{tab:lidarvsradar}
\begin{tabular}{c|ccc|ccc|ccc}
\toprule[.05cm]
\multirow{2}{*}{Sensor} & \multicolumn{3}{c|}{SS-night}& \multicolumn{3}{c|}{SS-moderate-rain} & \multicolumn{3}{c}{SS-heavy-rain}\\
\cline{2-4} \cline{5-7} \cline{8-10} \rule{0pt}{8pt}
 & R@1 & R@5 & R@10 & R@1 & R@5 & R@10 & R@1 & R@5 & R@10\\ 
\hline \rule{0pt}{8pt}
LiDAR & \textbf{94.4} & \textbf{96.9} & \textbf{97.1} & \textbf{93.7} & 94.5 & 96.8 & 87.7 & 90.3 & 94.2\\
4D Radar & 88.4 & 93.3 & 95.9 & 90.1 & \textbf{95.7} & \textbf{97.4} & \textbf{88.1} & \textbf{93.5} & \textbf{96.0}\\ 
\toprule[.05cm]
\end{tabular}
\vspace{-0.7cm}
\end{table}

\section{CONCLUSIONS} 
We propose a novel KD framework named 4DRaL that is designed to enhance 4D radar place recognition. By adopting knowledge distillation, 4DRaL boosts R2R performance and enables R2L place recognition. 
Extensive experiments validate that our method achieves state-of-the-art performance on both R2R and R2L tasks. This work lays a strong foundation for further advancements in 4D radar place recognition.

\bibliographystyle{ieeetr}
\bibliography{ref}

 \end{document}